 \documentclass[final,3p,times,twocolumn]{elsarticle}

\usepackage{amsmath,array,graphicx}
\usepackage[table]{xcolor}
\usepackage{multirow,multicol}

\usepackage{caption}
\usepackage{tabulary}
\usepackage[para]{threeparttable}
\usepackage{array,booktabs,longtable,tabularx}
\newcolumntype{L}{>{\raggedright\arraybackslash}X}
\usepackage{ltablex}
\usepackage{siunitx}
\usepackage{caption}
\usepackage[flushleft]{threeparttablex}

\usepackage[hyphens]{url}
\usepackage{hyperref}
\hypersetup{breaklinks=true}

\usepackage{amssymb}
\usepackage{xcolor}
\usepackage{titlesec}

\newcounter{bla}

\usepackage{graphicx}
\graphicspath{ {./figures/} }

\begin{document}

\begin{frontmatter}

\title {Task-oriented Dialogue Systems: performance vs. quality-optima, a review}

\author[a]{Ryan Fellows\corref{author}}
\author[a]{Hisham Ihshaish}
\author[a]{Steve Battle}
\author[a]{Ciaran Haines}
\author[a,b]{Peter Mayhew}
\author[a,c]{J. Ignacio Deza}




\cortext[author] {Corresponding author.\\\textit{E-mail address:} ryan.fellows@uwe.ac.uk}
\address[a]{Computer Science Research Centre (CSRC),\\ University of the West of England (UWE), Bristol, United Kingdom}
\address[b]{GE Aviation, Cheltenham, United Kingdom}
\address[c]{Universidad Atl\'antida Argentina, Mar del Plata, Argentina}

\begin{abstract}

Task-oriented dialogue systems (TODS) are continuing to rise in popularity as various industries find ways to effectively harness their capabilities, saving both time and money. However, even state-of-the-art TODS are not yet reaching their full potential. TODS typically have a primary design focus on completing the task at hand, so the metric of task-resolution should take priority. Other conversational quality attributes that may point to the success, or otherwise, of the dialogue, may be ignored. This can cause interactions between human and dialogue system that leave the user dissatisfied or frustrated. This paper explores the literature on evaluative frameworks of dialogue systems and the role of conversational quality attributes in dialogue systems, looking at if, how, and where they are utilised, and examining their correlation with the performance of the dialogue system.

\end{abstract}


\end{frontmatter}




\section{Introduction}\label{sec:1}

The field of dialogue systems can be split into two categories \cite{Su2016}; the sub-field of chat-oriented dialogue systems, which have the objective of relaying contextually appropriate and stimulating responses \cite{Vinyals2015}, and task-oriented dialogue systems (TODS), which are designed to assist a user in completing their goals. Examples include finding transport times, booking tickets or finding products \cite{henderson2014second}. 
There has been significant uptake in the adoption of TODS over recent years as companies are recognising their potential in alleviating the resource requirements inherent in human-based dialogue services. A prediction by market research firm Grand View Research estimates that the global chatbot market will reach \$1.23 billion by 2025 \cite{GVR,wang2018trust}. 
The literature exploring TODS performance generally focuses on bench-marking against human-generated supervised feedback, such as that of task-resolution \cite{walker1997paradise, Moller2006}; a measure that encapsulates the dialogue system's success rate in resolving a task or set of tasks. A direct correlation is assumed between task resolution and the performance of the dialogue system as a whole.  

Whilst task-resolution is priority, there are additional gains to be had in terms of performance and user experience as the journey of the whole conversation is considered, rather than just the destination of the task in question.

In addition to the task-resolution, within TODS performance evaluation studies, more in particular compared to other types of dialogue systems, user satisfaction is commonly considered as another performance metric, as an indicator of system efficiency\cite{williams-etal-2017-hybrid,Kamm10031} or  usability\cite{Moller2006}. The user satisfaction metric assumes a relative usability or efficiency for a dialogue system on the the basis of how its users are satisfied. These are usually approximated by two approaches: either by means of laboratory experiments, eliciting human judgment on system outputs and  behaviour relative to a predefined set of interaction parameters (eg. number of turns\cite{walker1998learning}, dialogue duration\cite{fraser79}). Or through modelling satisfaction, whereby the aim is to create models that provide ratings of performance similar to those which humans would do. The ratings based on human judgment are then used as target labels to learn an evaluation model based on objectively measurable performance attributes \cite{survey2020}.

Comparing the performance of dialogue systems is a non-trivial task, furthermore. This is due to the wide range of domains in which the systems are deployed, and the criteria they are evaluated against. Interactions are also subjective. What might be an optimal response for one individual, could be completely unsuitable for another, with performance being gauged on that specific individual's communicative preferences. 

This paper explores quality attributes that describe different qualities of conversational interactions between a system and the user, besides task outcome. We analyse conversational quality attributes in TODS and explore how they are utilised, and to what effect. To accomplish this, a literature survey is undertaken to examine current considerations to conversational quality attributes used in conjunction with dialogue systems, as well as frameworks to evaluate their performance as a multivariate function of multiple conversational quality attributes. \par

The rest of the paper is organised as follows: Section 2 explores TODS conversational quality attributes, and surveys their application to study and evaluate dialogue systems. Section 3 explores existing evaluation framework theory and application. Further discussion and conclusions are provided Section 4.

\section{Conversational Quality Attributes}
In a real world, human-to-human, task-oriented interaction, a conversation would likely not be deemed successful if only the task was resolved. If the advisor, in this situation, was friendly, personable and efficient in their manner, the advisee would be significantly more likely to have a positive experience and return in the future. However, if the advisor was rude or did not convey information competently, the advisee would most likely be left frustrated or even angry, leaving with a bad impression. Of course interactions with a human do not translate perfectly to interactions with machines, yet findings from real world communication can be extrapolated and applied to virtual communication. \par

In most circumstances, a TODS should elicit a positive user experience while seeking to resolve tasks in the most effective way possible. Accordingly, the evaluation of TODS performance generally seeks to optimise two main qualities: task-resolution and dialogue efficiency.

This section surveys the state-of-the-art developments on conversational quality attributes in the context of TODS, and highlights some of the most prominent attributes addressed in the literature around TODS performance.

\subsection{Task Resolution}
Task-resolution, or goal completion, is one of the most accessible metrics — and arguably can be the easiest to derive given a well-defined user goal as well as a predefined function to quantify a resolved, unresolved or somewhere between, task — to evaluate the success of a TODS. The main purpose of a TODS is to assist a user with a specific task in an automated fashion. Therefore, the success of a dialogue system in fulfilling information requirements established by user goals is an indicator of dialogue system's performance.

Practically, task-resolution (or success) is used to test dialogue systems success in providing not only the correct information, but also all user requested information —addressing as such the components for a given user-task: a set of constraints (target information, or information scope) and a set of requests (all required information)\cite{schatzmann2007agenda}. This in fact is consistent with the established understanding in Psychology around the notion of ‘conversation', that is, it is understood that when individuals engage in conversation, there is a mutual understanding of the goals, roles and behaviours that can be expected from the interaction \cite{clark1991grounding, clark1996using, clark1987collaborating, clark1989contributing}. Therefore, the ‘performance’ of the dialogue has to be evaluated on the basis of their mutual understanding and expectations.

In its simplest form, however, this metric can be quantified as a Boolean —binary task success (BTS)— value indicating whether a task or set of tasks has been resolved or not.  Using this metric, organisations can capture useful statistics over a number of interactions to derive how effective their dialogue system is at solving tasks, in comparison to interactions with human assistance or even other dialogue systems.

One of the more inherent challenges of task-resolution, as a performance metric, is knowing whether the task in question has been resolved. Especially so as the different users may have different goals, or intrinsically multiple goals, and these may even change in response to system behaviour throughout the course of interaction.  On top of this, different users may have varying  definitions of success, for example, a domain-specific expert user may deem a task resolved with less detailed information acquired compared to a novice user.

Typically an interaction with a dialogue system  will end when a user terminates the conversation, however this doesn't necessarily imply that their goals have been met. Some dialogue systems opt to explicitly elicit “task completion” in some form: \textit{has your request been resolved?} or \textit{is there anything else I can help you with?}, others attempt to use some form of classifier to infer when a task has been resolved through a machine learning and NLP model (eg. \cite{cambridge2015, classification15}). This requires a structured definition of goals and a mechanism to measure success relative to that goal. In this fashion, much of the work on automating the evaluation of task success has largely focused on the domain-specific TODS. This is usually an easier task as such systems can be highly scripted, and task success can be specifically defined – especially so in traditional dialogue systems, such as the Cambridge Restaurant System \cite{THOMSON2010562} and the ELVIS email assistant \cite{walker1998learning} — where the relevant ontology defines intents, slots and values for each slot of the domain.

However, a structured definition of goals will usually bind dialogue systems to a specific class of goals, constraining their ability to adapt to the diversity and dynamics of goals pertinent in human-human dialogue \cite{noseworthy-2017}. To address the shortcomings in adaptability and transferability encountered in single-domain systems, research into domain-aware, or multi-domain, dialogue systems has attracted noticeable attention in recent years \cite{multidomain2013, huang-etal-2020-meta, Wu2019TransferableMS}. This saw the introduction of the concept of domain state tracker (DST), which accumulates the input of the turn along with the dialogue history to extract a \textit{belief} state;  user goals/intentions expressed during the course of conversation. User intentions are then encoded as a discrete set of dialogue states, i.e., a set of slots and their corresponding values, e.g., \cite{mrksic-etal-2017-neural, statetracking2018, zhong-etal-2018-global}. As a result, the multiple user intentions are subsequently evaluated, whether objectively met or otherwise — reader is advised to refer to Figure 1 in \cite{Wu2019TransferableMS} for detailed characterisation of DSTs.

Reinforcement learning systems aim to find the optimal action that an automated agent can take in any given circumstance, by either maximizing a reward function or minimizing a cost function. With a dialogue system as the agent, the given circumstance is the belief state held by the DST, the reward function is linked to task-resolution, and the actions are the system's output slots and values. Dialogue systems will inevitably encounter problems; examples include incorrectly identifying a word, or a user changing their goal. A system could assign confidence levels to the belief states, track multiple belief states, and include a plan to recover the conversational thread after the errors are noticed.  

Casting the conversation as a partially observable Markov decision process (POMDP) allows for these uncertainties to be encoded \cite{williams2007partially, gavsic2013gaussian}. A POMDP is defined as a tuple $\{S, A, \tau, R, O, Z, \lambda, b_0\}$ where $S$ is a set of states describing the environment; $A$ is a set of actions that may be taken by the agent; $\tau$ is the transition probability $P(s^{\prime} \mid s, a)$; $R$ defines the expected reward $r(s, a)$; $O$ is a set of verifiable observations the agent can receive about the world; $Z$ defines an observation probability, $P(o^{\prime}  \mid s{\prime}, a)$; $\lambda$ is a geometric discount factor $0 \le \lambda \le 1$; and $b_0$ is an initial belief state $b_0(s)$. A POMDP dialogue system tracks multiple parallel belief states, selecting actions based on the belief state that is most likely. When misunderstandings occur, the current belief state can be made less likely, allowing the system to move to a new belief state. Because the belief states' probabilities are tracked alongside the expected action rewards and the chance that an action will transition as expected, a POMDP is able to effectively plan how to manage a dialogue. This framework allows a TODS to track multiple possible user goals, to plan error checking of user utterances, and to use context to potentially identify when the dialogue system has misunderstood the user intention. However, converting this potential benefit into practice is not trivial. Such systems are known to require a significant amount of training, as the state - action space can be very large even for single domains, and uncertainty in the task resolution may weaken the agent's learning \cite{zhang2020recent}.

In general, task-resolution is commonly quantified as the result of a performance metric in which user satisfaction is maximised. In PARADISE framework \cite{walker1997paradise} —reported later in Section \ref{sec:paradise}—, which is frequently used as a baseline for task success evaluation throughout literature, this is usually solved for user satisfaction as a weighted linear combination of task success measures side by side with dialogue costs (reported in Sec. \ref{dia:cost}). These measures can be objective, which entail features such as word error rate \cite{allen1996robust}, automatic speech recognition (ASR), word-level confidence score\cite{meena2014data}, number of errors made by the speech recognizer \cite{callejas2008relations} and time to fire, task completion rate, and accuracy metrics as used in \cite{robinson2007evaluation}, or subjective such as intelligibility of synthesized speech \cite{callejas2008relations} and perception tests \cite{meena2014data}.

\subsection{Usability and Dialogue Efficiency}

Usability attributes, such as user satisfaction, learnability, efficiency, etc, are the foundation of the design of ``successful'' dialogue systems, as these are ultimately created for the user, and for the user to achieve their intended, and occasionally variable, goal(s). While such attributes should ultimately be the criteria to evaluate a dialogue system, they are well-known to be subjective, and subsequently hard to measure. This is why much literature on evaluating dialogue systems tends to deal with quantifiable performance metrics, like task-resolution rate or elapsed-time —or turns— on task. It has been proposed, however, that an agent's competence in objectively measurable dialogue-quality attributes does not necessarily induce a better user experience, and subsequently a better system's overall usability \cite{Lamel2001}. In fact the different metrics may even prompt contentious interpretations, or simply contradict each other \cite{Kamm99}.

Although usability ratings are notoriously hard to interpret, especially if the system is not equipped to infer and keep track of user goals, the successful encapsulation of such values can provide insight that explicit metrics stumble to capture. From the study of Malchanau et al, usability experts rated examined questions from a 110 item questionnaire and derived an evaluation of their agreement of usability concepts. This led to a collection of 8 attributes they saw as key factors: task completion and quality, robustness, learnability, flexibility, likeability, ease of use and usefulness (value) of an application \cite{malchanau2019multimodal}. This questionnaire was used to evaluate a dialogue system designed for training purposes, in which the overall system usability was determined by the quality of agreements reached, by the robustness and flexibility of the interaction, and by the
quality of system responses.

Additionally, these different metrics may in fact have an inconsistent statistical interpretation to different designers. In the same way human evaluation will provide different outcomes based on the subjective criteria, the same can be said for metrics of usability which are difficult to consistently quantify \cite{raita2011too}.      

\subsubsection{User Sentiment}

Because of the insights sentiment analysis reveals about the more concise bodies of text on  social  media,  the field  of  sentiment analysis  has  seen  a take-up of use over recent times\cite{survey_opinionmining}. These can be performed on large quantities of tweets and posts from different platforms to assess general opinion about a specific product or topic. 

Different applications use a range of machine learning classification algorithms to categorise sentiment scores \cite{sentiment1,sentiment2}, some use just two classes: positive and negative, while others use an n-point scale, e.g., very good, good, satisfactory, bad, very bad \cite{prabowo2009sentiment}. A review and a comparative study of existing techniques for opinion mining like machine learning and lexicon-based approaches is provided in \cite{sentimenttwitter}.

Early studies on sentiment analysis in the context of dialogue systems explored the inclusion of user sentiment in rule-based systems, towards adaptive spoken dialogue systems, eg., \cite{emotion_sentiment, emotion_sentiment2}. Most of these studies investigated modular-based dialogue systems (conventionally referred to as pipeline models), with predefined rules for systems to adapt to variability in user sentiment.  In recent studies, however, much focus has been placed onto sentiment-adaptive end-to-end dialogue systems, particularly due to their adaptability in comparison with modular-based ones \cite{liu-lane-2018-end}, which are known to be harder to train, and adapt to new contexts \cite{braunschweiler18}.

Studies exploring the conjunction of dialogue systems with sentiment analysis are often motivated by the notion of system \textit{adaptability}, assuming a correlation between adaptability of the systems to user sentiment and their satisfaction. Some recent work emphasises the importance for conversational agents to adapt to different user (personality) types \cite{Liao2020,personality21}. Attention is paid to  studying user sentiment as a variable to guide the design of sentiment-adaptive dialogue systems \cite{sentiment-adaptive18, sentiment-adaptive20}. A comprehensive list of development milestones on sentiment analysis application to the analysis and evaluation  of dialogue systems, as well as on sentiment-adaptive systems is provided in Table\ref{tab:sentiment}.

\begin{table*}[!th]

\footnotesize
\caption{Main user sentiment studies in dialogue systems reviewed in the literature.}
\centering
\begin{tabular}{p{3cm}p{3.5cm}p{2cm}p{7cm}}
\hline\hline
Domain & Author & Year & Proposal / findings \\ [0.5ex] 
\hline

SDS& Schuller \cite{emotionSchuller2003}, Nwe \cite{emotionNwe2003}&2003& Emotion recognition in spoken dialogue using phonic features.\\  

SDS and TOSS &Devillers \cite{emotionDetection2003}, Liscombe \cite{Liscombe2005}&2003/05&  Automatic and 'robust' cues for emotion detection using extra linguistic features, lexical and discourse context. \\

DS & TH Bui \cite{MDP2006} & 2006 & \textit{'Affective'} dialogue model: inferring user's emotional state for an adaptive system's response. Earlier work applied to spoken dialogue systems in \cite{MDP_2000_spoken}. \\

SDS & Acosta \cite{emotion_sentiment, ACOSTA20111137} & 2010/11& \footnotesize{Gracie: inference of emotional state from utterance-by-utterance, and adaptive 'emotional coloring' system response. } \\

TODS and SDS &Ferreira \cite{Ferreira13}, Ultes \cite{Ultes17}& 2013/17 & Proposed an expert-based reward shaping approach in dialogue management, and a live user satisfaction estimation model based on 'Interaction Quality', a "less subjective variant of user satisfaction".   \\

DS & Shi\cite{sentiment-adaptive18}&2018& Detecting user sentiment from multimodal channels (acoustic, dialogic and textual) and incorporating the detected sentiment as feedback into adaptive end-to-end DS. \\

DS & Jaques\cite{jaques2020} & 2019 & Deep reinforcement learning model (off-policy batch RL algorithm).\\

DS & Shin\cite{shin2019happybot}&2019& Happybot: on-policy learning in conjunction with a user-sentiment approximator to improve a seq2seq dialogue model.\\

DS & Sasha \cite{Saha2020}&2020& Applying Reinforced Learning to manage multi-intent conversations with sentiment based immediate rewards. \\

\hline
\end{tabular}
\label{tab:sentiment}
\footnotesize{\textbf{DS}: Dialogue Systems, \textbf{SDS}: Spoken Dialogue Systems, \textbf{TODS}: Task-oriented Spoken Systems, \textbf{TOSS}: Task-oriented Spoken Systems.}\\
\end{table*}

It should be noted, nonetheless, that sentiment analysis methods have not been been  extensively applied to conversational agents and dialogue systems. One reason for this is the fact sentiment analysis performs more effectively when pre-trained on a domain specific dataset, and would not often generalise to open domains of discourse inherent in many dialogue systems. One example is the well-known shortcomings when generalising sentiment classification  of models trained on the IMDB movie database to classify sentiment about movies \cite{kumar2019sentiment, yenter2017deep, shaukat2020sentiment}. \par

However, as data becomes more accessible and the sentiment analysis techniques become more mature, the performance and scalability of many sentiment analysis tools are constantly improving. This in fact can allow for further advances in the development of \textit{sentiment-aware} dialogue systems, such that dialogue systems can adapt to the dynamics of user sentiment throughout the course of interaction. Depending on the objective function to optimise, there can be multiple approaches to extract and use the variability in user-sentiment, which can be categorised into two groups:  

\begin{itemize}
    \item \textit{Individual user utterance}: looking at the sentiment score of individual user utterance, which can offer insight into the specific semantics and vectors of that single interaction such as that found in \cite{agarwal2011sentiment, saif2012semantic}. This compartmentalised approach allows a deeper inspection of the content of that one message, whether this be a product, experience or other entity.
    \item \textit{Contextual user utterance}: the thread as a whole can be inspected from a temporal perspective, evaluating the evolution of the thread, rather than just individual messages \cite{fukuhara2007understanding, preethi2015temporal}. This can give insight as to why the sentiment of the user is going up or down and allows inspection as to why this is happening. When compared with other threads, trends can be found as to what is causing the fluctuation of sentiment. The difference of sentiment score between the first and last message, which can be referred to as the 'sentiment swing', can also be very useful, as this is an example of how the situation has progressed from the perspective of the user.
\end{itemize}

\begin{figure}[h!]
  \centering
  \includegraphics[width=0.45\textwidth]{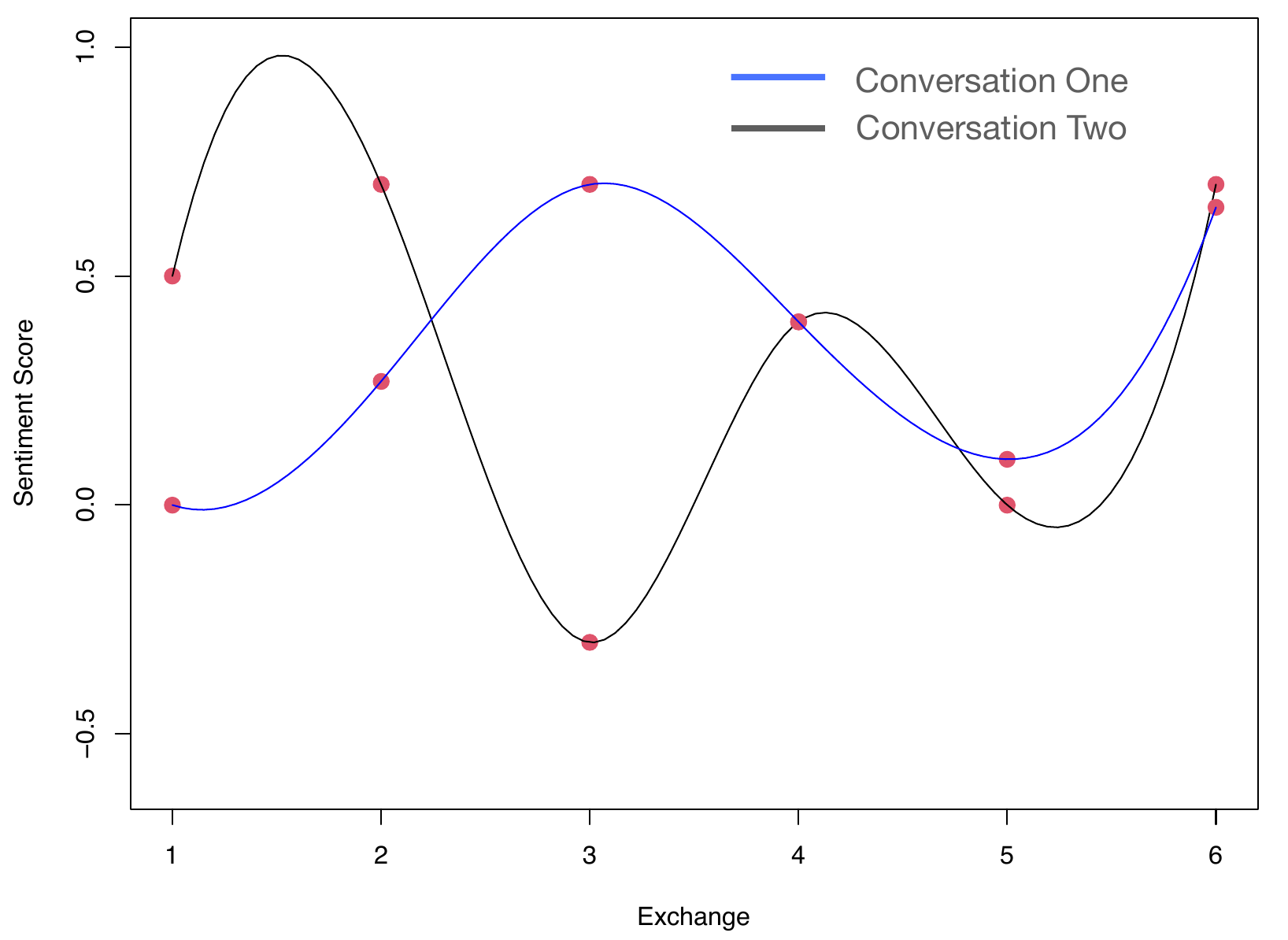}
\caption{Sentiment score of conversation one and two over the course of an interaction. }

  \label{fig:SAcam}
\end{figure}

An example to illustrate user-sentiment swing during dialogue is provided in Table \ref{tab}, which shows two "resolved" task-oriented interactions of 6 turns each. The sentiment score corresponding to user utterance at each turn is recorded. For simplicity, the variability in user-sentiment at each turn is smoothed in Fig.\ref{fig:SAcam}. Despite the similarity of their destinations, the two interactions exhibit diverging user-sentiment paths. While the first conversation has a positive swing of 0.7 over 6 turns,  eliciting a noticeable improvement over the entire interaction, the second stays nearly constant which could also be deemed positive, nonetheless. A substantially negative sentiment swing might be cause for examination to assess when considering designs for sentiment-adaptive dialogue.

\begin{table*}[!h]
\footnotesize
\begin{tabular}{clp{12cm}cc} 
\toprule
\multicolumn{2}{c}{} \textbf{Source} & \textbf{Utterance} &\textbf{Score} & Swing \\
\multirow{20}{*}{\rotatebox{90}{\textbf{Conversation 1}}}
&User & i need a place to dine in the center thats expensive & 0 \\
    \cmidrule(l){2-4}
     &Bot & I  have  several  options  for  you;  do  you  prefer African, Asian, or British food? & \\
     &User & Any  sort  of  food  would  be  fine,  as  long  as it is a bit expensive.  Could I get the phone number for your recommendation? &0.27 & $\uparrow$ \\
    \cmidrule(l){2-4}
    &Bot&There  is  an  Afrian  place  named  Bedouin  in the centre. How does that sound?&\\
    &User&Sounds good, could I get that phone number? Also, could you recommend me an expensive hotel?& 0.72 &$\uparrow$\\
    \cmidrule(l){2-4}
    &Bot&Bedouin's phone is 01223367660. As far as hotels go, I recommend the University Arms Hotel in the center of town.&\\
    &User&Sounds good, could I get that phone number? Also, could you recommend me an expensive hotel?& 0.4 &\color{red}$\downarrow$\\
     \cmidrule(l){2-4}
    &Bot&Sure, when would you like that reservation?&\\
    &User&i want to book it for 2 people and 2 nights starting from saturday.& 0.08 &\color{red}$\downarrow$\\ 
    \cmidrule(l){2-4}
    &Bot&Your booking was successful. Your reference number is FRGZWQL2. May I help you further?&\\
    &User&That is all I need to know. Thanks, good bye.& 0.7 &$\uparrow$\\ 
    &Bot&Thank you so much for Cambridge Town Info centre. Have a great day!&\\
    \midrule

\multirow{15}{*}{\rotatebox{90}{\textbf{Conversation 2}}}

  &User & Guten Tag, I am staying overnight in Cambridge and need a place to sleep. I need free parking and internet.&0.51 \\
\cmidrule(l){2-4}
     &Bot & I have 4 different options for you. I have two cheaper guesthouses and two expensive hotels. Do you have a preference? & \\
     &User & No, but I'd really like to be on the south end of the city. Do any of those fit the bill? &0.75 & $\uparrow$ \\
    \cmidrule(l){2-4}
    &Bot&Sure. Does price matter? We can narrow it down and find exactly what you need.&\\
    &User&No I don't care about the price. Which one do you recommend?& -0.32 &\color{red}$\downarrow$\\
    \cmidrule(l){2-4}
    &Bot&I would recommend aylesbray lodge guest house. Would you like me to book that for you?&\\
    &User&Yes, book it for 4 people and 4 nights starting from tuesday.& 0.4 &$\uparrow$\\
     \cmidrule(l){2-4}
    &Bot&The booking was unsuccessful. would you like another day or a shorter stay?&\\
    &User&How about for 3 nights?& 0 &\color{red}$\downarrow$\\ 
    \cmidrule(l){2-4}
    &Bot&Booked! Reference number is: 84ESP6F5 &\\
    &User&Great. I am all set then. Have a nice day. Bye.& 0.78 &$\uparrow$\\ 
    &Bot&Have a nice stay. Bye.&\\
    \bottomrule
\end{tabular}
\caption{\footnotesize{Sentiment score of conversation one and two over the course of an interaction. The conversations are from the Camrest dataset \cite{wen2016network}, with the sentiment scores being generated with VADER sentiment compound \cite{VADER}, as an an aggregated value of the positive, negative and neutral score.}}
\label{tab} 

\end{table*}

\vspace{1cm}

Despite the scores fluctuating throughout the interaction, both threads end with a positive conclusion, indicating that the user was satisfied with the outcome. Whilst this is insightful in itself, the highs and lows provide a chance to understand why these values were exhibited at that point, which could allow for the examination of the objective attributes or semantics used, or that the values could just simply be the result of a contextual issue, such as, in this case, a restaurant being fully booked.

However, regardless of the domain in which sentiment analysis is utilised, a cautious apprehension should be taken in interpreting the obtained scores. Modern sentiment analysis tools are advancing, but they are still not mature enough to accurately recognise sarcasm, jokes and nuances of language. There is also the limitation of a lack of distinctive sentiment annotations amongst an already limited amount of datasets readily available, as observed in \cite{saif2013evaluation}, which subsequently makes it harder to perform accurate analysis on dialogues of a more extensive lexicon.  

What's more, sentiment analysis is sensitive to social conventions which are prevalent in human communication. Many interactions through email, for example, will exhibit some form of generic greeting such as 'Good Morning' as well as a sign off (sometimes inserted automatically through a template) such as 'Best Wishes'. These terms are often used by individuals, regardless of the context of their email, which can therefore skew the sentiment score to be higher than the actual substance of their email might elicit. 

Therefore it could be argued that the current state of sentiment analysis makes it a useful tool to gain analytical insight from a corpus of text, but to utilise them as the sole driver for action could potentially lead to erroneous decision making. The context of its usage is important.

\subsubsection{Dialogue Cost}
\label{dia:cost}
The term ‘dialogue cost’ appears frequently throughout dialogue system literature \cite{scheffler2002automatic, walker1997paradise, relano1999robust} and typically refers to multiple aspects of resource retrieval and utilisation ranging from the data itself, to the computational power required by the model being utilised. Some literature even refers to  the  explicit  monetary  cost  of  the  dialogue  system based on the manual labour required to label the data, often using the method of crowdsourcing \cite{mitchell2014crowdsourcing, shah18}.

Relevant and feature rich data is the foundation for a high  performing  dialogue  system,  and  no  matter  how good  a  model  is,  it  cannot  compensate  for  a  small or poor quality dataset. Therefore such resources can be expensive to acquire, whether in terms of time or money \cite{manuvinakurike2015reducing}. In more domain specific dialogue, the data native to these sometimes unfamiliar domains, plays an even more important role as it highlights semantic and pragmatic phenomena that is unique to that domain. 

Alongside task-resolution, dialogue cost is often  considered to  infer  ‘dialogue strategies’ \cite{levin1997learning, scheffler2002automatic}, which specify at each stage what the next action to be taken by the system. A dialogue strategy can have the objective of converging towards the goal state in the most efficient way possible through a series of interactions with the user. ‘Efficiency’ can for example, mean access to external resources, the dialogue duration,  internal computation time, or resource use. The goal is to reduce these ‘costs’ to allow the system to achieve higher performance.

However, the ambiguity of the term ‘dialogue cost’ can make it a difficult area to assess. The PARADISE framework (see Sec.\ref{sec:paradise}) describes efficiency measures such as the number of turns or elapsed time to complete a task \cite{abella1996development, hirschman1993cost, walker1997paradise}, as well as qualitative measures such as inappropriate or repair utterances \cite{danieli1995metrics, simpson1993black} as potential dialogue costs. Whereas, some researchers explore the term from a reinforcement learning perspective, in which the dialogue cost is a penalisation assigned for taking the wrong action predicated on a pre-defined function. Therefore, it can be a difficult to quantify cost in relation to a dialogue. Even when considering what is typically agreed on, regardless of the context, that dialogue 'cost' should be minimised, i.e., to maximise system efficiency, there isn't such established foundation to suggest that, for instance, a shorter —hence more ‘efficient’– dialogue is directly correlated to a better user experience. In fact it can simply be the opposite.

\subsubsection{Retention Rate}
The retention rate of a TODS is often refereed to as a measure of the number users that return to use the system within a given time frame. This is another important, yet accessible metric for quantifying dialogue systems' performance.  If a company’s  chatbot  aims  to  replace  other  communication channels  (e.g.,  lowering  call  volume),  the  goal  is  to obtain  significantly  higher  retention,  which  can be indicative of higher consumer satisfaction \cite{Dhyani2019}. However,  there  are  plenty  of  other  automated options that allow users to manage accounts easily without speaking to a human. Thus, if a chatbot is focused on customer support, a high retention rate does not necessarily have to be the measure of success \cite{przegalinska2019bot}.

The context and domain in which the TODS is deployed is an important factor to consider when looking at the retention rate of a given dialogue system.  If the dialogue system in question is a health-based chatbot  for  a  one-off issue,  then  the  user  is unlikely  to  have  to  reuse  the  chatbot,  and  therefore the metric is less valuable. However, if the chatbot is being deployed as a customer service replacement, then a high retention rate can be interpreted as a positive performance indicator, as it shows the user has enough confidence in the system to reuse it. 

Related metrics are those of \textit{dropout} rate and \textit{bounce} rate. The dropout rate refers to the number of users who quit the session with the dialogue system before an outcome had been reached. A high dropout rate for a dialogue system can be a substantial indication of poor performance. The bounce rate is the volume of users that do not utilise the dialogue system for its intended use. A high retention rate with low dropout and bounce rates would suggest a high level of performance. 

However, only so much can be derived from the metric of retention rate without some form of user feedback, as the metric is sensitive to anomalies. A dialogue system could perform perfectly, yet a user might not return for other, unknown reasons. This should not be indicative of the performance of the system, yet the metric might suggest this to be the case. Therefore, the larger the set of interactions retention is analysed on, the more insightful the findings will potentially be. Because of this, it could be argued that the rate of retention offers a good overview perspective of system performance, but such considerations should prevent retention rate from being a primary form of performance insight.  

\subsubsection{Response Time}
The literature exploring dialogue response time is typically concerned with reducing the time it takes a conversational agent to respond to the user. The consensus is that a user wants responses as quickly as possible, and for the interaction to be as efficient as possible in terms of session time. The focus is often on the mechanics of the model in question, rather than the effect that  response time could have on user satisfaction \cite{nursetyo2018smart, Kowsher2020}.
Alternative  studies  on response  time  shift  the  focus  from  the desire for  instant  responses to adding more human-like delays. In their study of using  dynamic response delays for machine generated messages, Gnewuch  et  al \cite{gnewuch2018faster} prioritise  the  `feel'  of  the  conversation over speed of response, opting to ``calculate a timing mechanism based on the complexity of the response and complexity of the previous message as a technique to increase the naturalness of the interaction''. As a result of these dynamic delays, they showed an increase in both the perception of humanness and social presence, as well as a greater satisfaction with the overall dialogue interaction; a faster response time is not necessarily better. 
However, as with the majority of the quality attributes, the context and domain are very important to consider. ‘Replika’ \cite{replika.com} is an anthropomorphised chatbot designed as a companion to help battle loneliness. It utilises a slight delay to make the interaction feel more genuine and human-like, as instant replies would make the interaction feel too machine-like and  break  the  social  illusion. Conversely, ‘911bot’ \cite{911bot} is a chatbot that allows a user to describe an emergency situation, and because of this context, any artificial delays would  not  be  appropriate. This highlights the importance of context when considering such conversational attributes to evaluate TODS performance. 

Computationally, response time has become much less of a pressing issue in recent times, as abundant computational resources, and innovation in machine learning \ NLP approaches, make instantaneous responses entirely feasible, and as a result, expected. Therefore, it could be argued that whilst a dialogue system might not get praised on its performance for optimal response times, whether instant or timed, it will be negatively graded for sub-optimal response times.

\subsubsection{Conversation Length}
The literature exploring the explicit length of conversation is limited.  This is due to the fact that the developers predominantly focus on the substance of a message first,  with the subsequent message length being as long or short as it needs to be.  However, the length of an agent's responses can significantly alter the dynamic of an interaction, as it determines how much information can be conveyed in a single turn. Depending on the topic at hand, if the messages are too short, there is a risk the user will grow frustrated with the lack of detail in the answer, but if the messages are too long, the user’s attention may wander. 

In their guide to developing ``better'' chatbots for mental health, Dosovitsky et al \cite{Dosovitsky2020} argue that ``developers should strive to find a module length that enhances intervention fidelity without compromising engagement'' and ``should focus on creating a few engaging and effective modules at the beginning rather than developing a large variety of untested modules''. Simply put,  system utterance length should be dynamic, changing relative to the stage of the conversation.

Other work examined the effect of message length relative to the dialogue domain, e.g., \cite{przegalinska2019bot}, emphasising that one of the most important chatbot performance metrics is conversation length and structure. Industry trends suggest aiming for shorter conversations  with  simple  structure, in line with the notion of efficient service. For example, banking chatbots  are assumed to  provide  quick  solutions such as sending and receiving money, or checking a balance. When the social aspect of the conversation is more important, fast and concise responses may turn counterproductive. 

However, just looking at conversation length from an objective perspective can be misleading. If an analysis is performed in which it is deemed shorter messages are preferred for a given domain, and are subsequently rewarded, then this may undermine the very relevant factor of context. Dialogue systems often have the objective of being as efficient as possible, which would encourage the idea of concise discourse, which may not be a problem. However, some issues and topics simply do not lend themselves to this approach and require further development in the conversation. Therefore, it would be detrimental to the system to simply penalise longer message without any thought to the semantics and context involved. This is not to say conversation length is not a useful quality attribute, as the literature suggests, it is, yet the optimisation of this parameter needs more than just the configuration of a value for utterance length or number of turns.

\section{Evaluation}\label{sec:3}

TODS are difficult to evaluate on a consistent basis. Comparing dialogue systems using human judgement can be problematic as this often varies  from one person to another, an issue that is pointed up by the fact that there is no current standard to measure against. Evaluating dialogue system performance relative to statistical criteria, such as task outcome or rate of retention, does allow for a consistent, though blunt, assessment. Besides, these methods often lack contextual information, which is a necessary component for in-depth evaluation of dialogue.

\subsection{Evaluation Methods}

The methods to evaluate dialogue system typically fall into three categories; human evaluation, user behaviour modelling and automatic evaluation. 

\subsubsection{Human Evaluation}

A comprehensive evaluation of a machine-produced dialogue interacting with humans eventually requires their judgment. Conventionally this has been carried out through bench-marking dialogue relative to human-generated supervised feedback, see \cite{walker1997paradise} and \cite{Moller2006}. In recent times,  the process of dialogue system evaluation by humans has become increasingly commonplace \cite{finch2020towards}. \textit{Evaluators} are often recruited via crowdsourcing to rate system generated responses relative to criteria such as ‘appropriateness’, ‘empathy’ and ‘helpfulness’ as used in \cite{xu2017new} or ‘grammar’, ‘context relevance’ and ‘correctness’ as used in \cite{zhu2017flexible}. Previous research which employs crowdsourced judgments has used metrics such ease of answering, information flow and coherence \cite{liu2016not}, ‘naturalness’ \cite{asghar2018affective} and ‘interestingness’ \cite{asghar2018affective, santhanam2019survey}.

Such process however is time consuming, and expensive. Arguably a more pressing issue is the lack of a standard for humans to use as a baseline, which can cause issues of consistency when evaluating dialogues \cite{eckert1997user}. 

To overcome the challenges arising from the subjective rating of humans, further approaches to the evaluation of dialogue systems' performance have been adopted, namely methods for user-modelling, and automatic evaluation. 

\subsubsection{User-modelling}

User-modelling, or simulation, aims at simulating users' interaction behaviour with the agent, which is used as a training environment for a dialogue system. The approach assumes that there is a small collection of annotated in-domain dialogues available \cite{schatzmann2007agenda}, or out-of-domain dialogues that have a matching dialogue format \cite{schatzmann2006survey}. In instances where there is no such data available, manually crafted values can be attributed to the simulation model arguments, given that the model is simple enough \cite{levin2000stochastic}. 

To achieve a seemingly realistic interaction through simulation, a significant amount of work needs to be done. For a high level virtual patient to be created, Campillos-Llanos et al \cite{campillos2020designing} demonstrated the need for the formalisation of ontological concepts for natural language understanding (NLU). NLU in fact could additionally rely on text requiring representations for resolving recurring paraphrasing concerns \cite{nirenburg2009integrating}, e.g., a collection of texts for questions and replies overseen by experts \cite{kenny2011embodied} and \textit{canned} questions and answers \cite{benedict2010virtual}.

\subsubsection{Automatic Evaluation}

The process of the automatic evaluation of dialogue systems does not require a direct human interference once the evaluation script has been written \cite{finch2020towards}. The aim is  to provide an objective measurement of the system performance by quantifying various attributes of dialogue into mathematical formulations \cite{finch2020towards}. By utilising automatic evaluation, the process of recruiting human \textit{evaluators}, which can cost significant time and money, can be avoided. 

BLEU \cite{papineni2002bleu} for example, is a metric that evaluates a generated sentence to a reference sentence, giving a score of 1.0 for a \textit{perfect} answer and 0.0 for a completely inappropriate answer. It is typically used to evaluate machine translation performance but has since been applied  in the field of dialogue systems \cite{chen2011amber, noorbehbahani2011automatic}. Another widely utilised evaluation metric is ROUGE, a method of text summarisation which uses measures to automatically determine the quality of a summary by comparing it to an ideal one created by a human \cite{lin2004rouge}. \par

Automated evaluation methods provide the most efficient and undemanding approach to assess dialogue systems \cite{finch2020towards}. However, they have also drawn criticism in the field. Liu et al \cite{liu2016not} argue that automated metrics are generally perceived as sub-standard indicators of true dialogue quality. Other studies suggest  that automated metrics can have a poor correlation with the judgements made by humans, e.g.,\cite{novikova2017we} , indicating poor utility of such metrics.


\subsection{Evaluation Frameworks}

\subsubsection{PARADISE Framework}
\label{sec:paradise}
The Paradigm for dialogue system evaluation, or PARADISE, is one of the earliest frameworks developed for the purpose of evaluating spoken dialogue agents and is still widely used as a baseline to evaluate dialogue systems' performance. The strengths of the framework include separation of task requirements from an agent's dialogue behaviours, the ability to compare competing dialogue strategies, as well as supporting the calculation of a metric of \textit{performance} spanning both sub-dialogues and whole dialogues \cite{walker1997paradise}. Because of the explicit metrics that can be generated from these methods, the framework can measure the capabilities of multiple agents performing different tasks, whilst allowing normalisation over task complexity \cite{walker2000towards}.

Researchers had previously used evaluation techniques, such as having a reference answer to compare against \cite{Hirschman1990BeyondCA}, or using various metrics to compare different dialogue strategies, for example, inappropriate turn correction ratio, concept accuracy, utterance ratio and implicit recovery \cite{Danieli1996MetricsFE, hirschman1993cost, polifroni1992experiments}. However, these proved insubstantial for the rapidly improving dialogue systems of the time. The PARADISE framework sought to overcome the limitations of these approaches and addressed the following three research goals\cite{walker1997paradise}:

\begin{itemize}
    \item ``To support the comparison of multiple systems or multiple versions of the same system doing the same domain tasks''.
    \item ``To provide a method for developing predictive models of the usability of a system as a function of a range of system properties''.
    \item ``To provide a technique for making generalizations across systems about which properties of a system impact usability, i.e to figure out what really matters to users''.
\end{itemize}

The framework gives designers the opportunity to predict user satisfaction, derived from a linear combination of objective metrics such as mean recognition score and task completion \cite{Kamm99, walker1997paradise, litman1999empirically}. It elicits a performance function that can be quantified with user satisfaction as the dependant variable, with task success, dialogue \textit{quality}, and dialogue \textit{efficiency} measures as independent variables \cite{ smeele2003evaluation}.

Practically, a task and a set of scenarios need to be initially defined. Table \ref{avm1tab}, used for illustration in \cite{walker1997paradise}, summarises the task information requirements of a train timetable service, through the representation of an attribute value matrix (AVM). Each attribute is coupled with a value which is obtained through the interaction between user and system. The information flow indicates who the information is being acquired by in that turn (agent or user), although this information is purely informative and not further used for evaluation.

\begin{table}[!th]

\footnotesize
\centering
\begin{tabular}{p{2cm}p{3.5cm}p{1cm}}
\hline\hline
Attribute & Possible Values & Information Flow \\ [0.5ex] 
\hline

depart-city  & Milano, Roma, Tornio, Trento   & to agent \\
arrival-city  & Milano, Roma, Tornio, Trento    & to agent \\
depart-range   & morning, evening  & to agent  \\
depart-time & 6am, 8am, 6pm, 8pm  & to user \\

\hline
\end{tabular}
\caption{An AVM of a simplified train timetable \cite{walker1997paradise}.}
\label{avm1tab}
\end{table}

Task success accordingly can be quantified as to how well the \textit{process} of achieving the task requirements have gone by the end of the interaction. A confusion matrix is utilised to represent the information requirements of a task for a set of dialogues, instantiating a set of scenarios \cite{walker1997paradise}.  This then is summarised by the Kappa coefficient $\kappa$ \cite{carletta1996assessing},  also referred to as Kappa statistic \cite{walker1998learning}, a quantity that is used to measure agreement after having corrected for chance,

\[\kappa =\frac{ P(A) - P(E)}{1 - P(E)} \]

$\kappa$ is derived from the confusion matrix values, as an objective measure to calculate a metric of agreement for the correct responses for each task scenario, where $P(A)$ is the ratio of agreements between the actual set of dialogue and AVMs, and $P(E)$ is the ratio of instances that the AVMs are expected to agree on. The measure is thought to be an appropriate form of dialogue system measurement, because it takes into account the inherent ambiguity of human – computer interaction by this correction for chance expected agreement \cite{walker1997paradise}. 

PARADISE was an innovative framework of its time, and whilst its evaluative strengths are comprehensive, it has its limitations. The utilisation of incorporating a user satisfaction score as a performance measurement can be powerful and harness a metric that is often overlooked in other dialogue evaluation techniques. However, Evanini et al \cite{evanini2008caller} raise the question of reliability with user satisfaction surveys, stating that individuals tend to have different interpretations of the question, which can cause issues of consistency when incorporated into the model. Ultes et al \cite{ultes2013quality} believed the act of users having to rate live dialogue is also an impractical one in a real-world environment.\par 

The framework is far less utilised for TODS evaluation in modern times, and is instead frequently used as a baseline reference point for newer evaluation techniques. The rigours and necessary characteristics of more \textit{sophisticated}  ‘chatbots’, arguably incorporate too many complexities for the PARADISE framework to effectively assess and therefore, practically, it is now used much less.

\subsubsection{Analytic Hierarchy Process}
Developed between 1971 and 1975, L Saaty designed and established the Analytic Hierarchy Process (AHP) as a general theory of measurement, which ‘can be utilised to derive ratio scales from both discrete and continuous paired comparisons’ \cite{saaty1987analytic}. Since its inception in the 1970’s, the framework has been the basis for further studies and developments, with variants such as the fuzzy AHP \cite{van1983fuzzy} and the Analytic Network Process (ANP) taking inspiration from the original concept. AHP’s have been adapted for a wide spectrum of domains ranging from agriculture \cite{alphonce1997application} to the military \cite{korkmaz2008analytic}, supplemented by the vast literature on the applications of AHP, with more than 1300 papers and 100 doctoral dissertations studying the topic \cite{forman2001analytic}. As explored in this section, the AHP can be utilised effectively to evaluate dialogue systems providing an interaction attributes are effectively represented.

\begin{figure}[h!]
\centering
\includegraphics[width=0.45\textwidth]{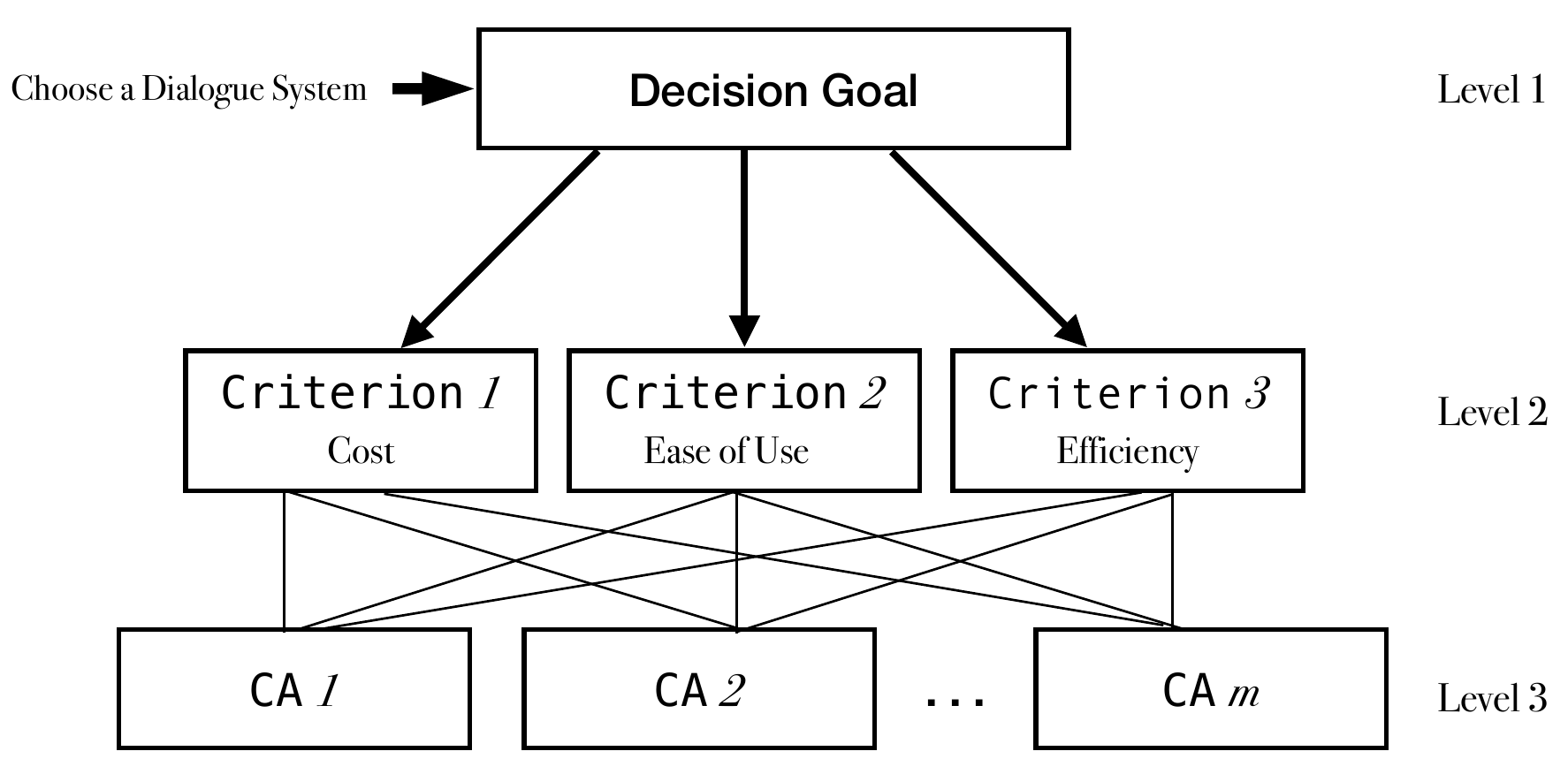}
\caption{A top down view of the AHP hierarchy. CA \textit{1}, \textit{2}, \textit{3}, ..., $m$ are alternative candidates, where CA is a conversational agent.}
\label{fig:ahp1}

\end{figure}

For an AHP to be effectively utilised, the problem in question needs to be modelled as a hierarchy. This hierarchy is made up of a goal at the top level, such as ‘choose a dialogue system’, with a number of relevant factors that link the potential candidates to the top level goal/question. The factors for the choice of dialogue system might be; efficiency of the system, ease of use of the system, standard of the systems user interface, and cost of the system, which in this case will refer to the monetary cost of implementation. Each of these factors will link to the three possible dialogue systems which will be the possible options/candidates, as shown in figure \ref{fig:ahp1}.

Nodes are represented in a AHP by a box. A node can be the goal, a criterion or a candidate. From the perspective of a given node, any nodes above that node are parent nodes, and any below are children nodes or sub-nodes.  Initially, the criteria need to be manually populated with the appropriate nodes, from which a series of pairwise comparisons can be made between attributes to determine their relative importance. If attribute A is substantially more important than attribute B when evaluating the pair against each other, then if A is rated at a 9, B should be rated at 1/9. A scale of 1-9 is typically used for AHP's. This is known as the Saaty rating scale \cite{saaty1990make}, as demonstrated in Table \ref{avm2tab}, which compares the conversational agents ease of use and gives a caption to justify these scores —an optional addition. Comparison matrices are derived and the first principal eigenvector of each one is computed to assess relative and global priority \cite{radziwill2017evaluating}, as shown in Table \ref{avm3tab}.

\begin{table}[!th]

\footnotesize
\centering
\begin{tabular}{p{0.6cm}p{0.1cm}p{0.6cm}p{0.1cm}p{4cm}}
\hline\hline

CA1  & 1  & CA2 &4 & CA2 offers a very intuitive user  interface, with a  lot of help ready if needed, CA1 is harder to use initially but does become easier after getting used to it. \\
\hline
CA 1  & 4  & CA 3 &1 & Neither possess ease of use as as strength, but CA1 gives the users prompts eventually. CA3 is almost impossible to use for a user who is not savvy with this technology already. \\
\hline
CA 2  & 9  & CA 3 &1 & is better in almost every way in comparison to CA3 for ease of use. A child would understand how to use CA2 instantly but would have no chance with CA3. \\
\hline

\end{tabular}
\caption{Options compared in relation to Ease of Use.}
\label{avm2tab}
\end{table}

\begin{table}[]
\centering
\small
\begin{tabular}{l|c|c|c|l|c|}
\cline{2-4}
\cline{6-6}
& CA1 & CA2 & CA3 &  & \begin{tabular}[c]{@{}c@{}}Priority \\ Vector\end{tabular} \\ 
\cline{1-4} 
\cline{6-6} 
\multicolumn{1}{|l|}{CA1} & 1            & 1/4          & 4            &  & 0.217    \\
\multicolumn{1}{|l|}{CA2} & 4            & 1            & 9            &  & 0.717  \\ 
\multicolumn{1}{|l|}{CA3} & 1/4          & 1/9          & 1            &  & 0.66   \\ 
\cline{1-4} 
\cline{6-6} 
\end{tabular}
\caption{Pairwise comparison matrix for ease of use, with corresponding priority vector/first principal eigenvectors.}
\label{avm3tab}
\end{table}

\vspace{1cm}
For the ‘choose a dialogue system’ example, a coefficient of priority needs to be assigned to each criterion on the process of choosing the correct dialogue system candidate. An equally important process is to determine the weight to be assigned to each candidate relative to each of the criteria, dependent on how important the people involved perceive it to be. The AHP is particularly useful in this respect because it allows the designers to input an explicit value to each of the criteria.

The priority coefficients, or weights of the nodes in any group are assigned priority values, which are absolute values between 0 and 1. If a node has a value of 0.3, it has twice as much chance of reaching the goal in comparison to a node with a value of 0.15. The value of the goal node is always 1, with the corresponding alternatives summing to 1. The weights of the children nodes will always sum to the priority value of the parent node. As represented in Table \ref{avm4tab}, the designers have to assign the criteria to the necessary values in respect of the goal, which is again done with pairwise comparisons.

\begin{table}[]
\small
\begin{tabular}{l|c|c|c|l|c|}
\cline{2-4} \cline{6-6}
\textbf{}                                 & EoU & Cost & Efficiency &  & \begin{tabular}[c]{@{}c@{}}Priority \\ Vector\end{tabular} \\ 
\cline{1-4} 
\cline{6-6} 
\multicolumn{1}{|l|}{EoU}        & 1            & 5             & 7                   &  & 0.726                                                               \\ 
\multicolumn{1}{|l|}{Cost}       & 1/5          & 1             & 3                   &  & 0.179                                                               \\ 
\multicolumn{1}{|l|}{Efficiency} & 1/7          & 1/3           & 1                   &  & 0.097                                                               \\ 
\cline{1-4} 
\cline{6-6} 
\end{tabular}
\caption{Pairwise comparisons of the criteria of the problem.}
\label{avm4tab}
\end{table}

Table \ref{avmtab5} shows the steps  required to synthesise the final priorities. The value for each criterion is multiplied by the priority vector for that attribute, which is done for each option. These values are collated in Table \ref{avmtab6} for illustration, with \textit{Conversation Agent 2} (CA2) having the highest goal score of 0.658, making it the best dialogue system using this criteria.

\begin{table}[!th]

\footnotesize
\centering
\begin{tabular}{p{1cm}p{1cm}p{0.6cm}p{1cm}p{1cm}p{0.6cm}}
\hline\hline
Criterion & P vs G & Alt. & A & B& C  \\  
\hline

&&CA1&0.217&&0.157\\
EoU&0.726 &CA2&0.717& x 0.726 & 0.52\\
&&CA3&\underline{0.066}&&\underline{0.047}\\

&&&1.000&&0.724\\

\hline

&&CA1&0.265&&0.01\\
Cost &0.179 &CA2&0.672& x 0.179 & 0.12\\
&&CA3&\underline{0.062}&&\underline{0.011}\\
&&&1.000&&0.141\\

\hline

&&CA1&0.743&&0.072\\
Efficiency&0.097 &CA2&0.194& x 0.097 & 0.018\\
&&CA3&\underline{0.063}&&\underline{0.006}\\
&&&1.000&&0.096\\

\hline

\end{tabular}
\caption{Table of all calculations for conversational agents results for each attribute. P vs G: Priority vs Goal; Alt.: Alternative CA.}
\label{avmtab5}
\end{table}

\begin{table}[!th]

\footnotesize
\centering
\begin{tabular}{p{1cm}p{1cm}p{0.6cm}p{1cm}p{1cm}p{0.6cm}}
\hline\hline
Candidate & EoU  & Cost & Efficiency & UI & Goal  \\  
\hline

CA1  & 0.119          & 0.024          & 0.201               & 0.015          & 0.359         \\
CA2  & 0.392          & 0.01           & 0.052               & 0.038          & 0.492 \\ 
CA3  & 0.036          & 0.093          & 0.017               & 0.004          & 0.15 \\
\hline
Total & \textbf{0.547} & \textbf{0.127} & \textbf{0.27}       & \textbf{0.057} & \textbf{1}    \\ 
\hline

\end{tabular}
\caption{Final results of each conversational agents score for each attribute, with totals.}
\label{avmtab6}
\end{table}

AHP’s are generally agreed to be easy to use in which the user(s) can \textit{plug in} their nodes and values following simple instruction. Just by doing this, a systematic outcome can be achieved which allows for consistency in decision making which is backed by a logical process. It also simplifies what can be a daunting decision-making process, into smaller, easy to digest steps \cite{kaliyamurthi2017comparison}. However, AHP’s still require human judgement to be quantified into numerical values, which can be subjective and hard to agree on amongst parties.

In these situations, it can be argued that the process is unnecessarily lengthened. The rigidness of the frameworks methodologies can make it hard to take into account uncertainty, a component often inherent in the domains they are often deployed in.

\subsubsection*{ChatEval}

ChatEval is a unified framework that harnesses already existing tools and provides a platform on the web for researchers to collaborate with their dialogue systems \cite{sedoc2019chateval}. The benefits of this tool are significant, as even though there is a pre-existing community in the NLP and dialogue system space, there is less opportunity for researchers and developers to compare their findings. This is one of the reasons the Loebner prize is almost held as a standard in itself \cite{powers1998total}, as it is a forum in which chatbots can be directly compared to others in the field. 

The fields of dialogue systems and conversational agents suffer from the issue of reproducibility, as replicating the exact threads of conversation can prove difficult, subsequently making it a tough task to assess exactly how a model is performing \cite{fokkens2013offspring}. This is an issue that is problematic for the field of NLP in general, and especially so in the research into dialogue systems in particular, because unlike other machine learning fields, there is a significant lack of \textit{automatic} metrics to evaluate dialogue system outputs.

Unlike the previous frameworks examined, ChatEval is not a standalone framework in its own right. The evaluation toolkit of ChatEval utilises both an automatic evaluation component and human evaluation component. The automatic evaluation consists of various components such as a BLEU-2 score over the mean of the sentence \cite{liu2016not}, a metric of mean cosine-similarity \cite{thongtan2019sentiment}, a lexical diversity metric \cite{Li2015} and response perplexity score \cite{Zhang2018}. The human evaluation segment involves a human choosing the better (or equal) response from a prompt and two possible responses. \par

Although the overall concept behind ChatEval is a simple one, it helps fill a void in the field, allowing for real progress to be made. In their machine learning problems, especially those of a classification nature, competitions can be held to see who can get the highest accuracy through  bench-marking for a given dataset, which encourages substantial variation in terms of the models utilised. ChatEval essentially offers the same opportunity for dialogue systems.

Whilst the bench-marking capabilities of ChatEval can offer substantial utility, the rigidity of the framework makes it difficult to evaluate interactions on the go. This limits the ability to optimise the conversation as they happen, and instead offers a forum of delayed feedback to allow for future design \textit{tweaks}. Therefore ChatEval can be of great use for dialogue system evaluation, but the context upon how it will be used holds significance.

\section{Discussion and Conclusions}

It is clear that there is no shortage of studies exploring the field of TODS and their performance. However, research into TODS in conjunction with conversational quality attributes, beyond that of task-resolution, are less abundant. One potential reason for this is because these attributes, such as conversation length, response time and user-sentiment are often referred to more as bi-products of the dialogue systems performance in meeting user information requirements.

Although many studies on optimising TODS performance examined metrics for performance evaluation beyond that of task-resolution, thus far, however, the modelling of TODS performance as a multivariate function of multiple conversational quality attributes remains an open question.

Additionally, TODS are still difficult to evaluate. Although there are established methods and frameworks which are frequently referred to in literature, with PARADISE arguably the most applied, yet there is still no standard in place for a novel TODS to be measured against. This is undoubtedly a hindrance to the field, as it gives a lack of consistency when designing a system and subsequently comparing it with others in the industry. Also, with the growing complexity of modern virtual assistants such as Siri, Bixby and Alexa to name a few, where each could be described as a \textit{sophisticated} TODS, the task of objectively evaluating such systems is only going to become a more complex process.

Therefore, although significant progress has been made in the field of TODS over a relatively short time, there are still various challenges to be overcome. Arguably the most pressing issue is the lack of a standardised protocol for human evaluation, which makes it challenging to compare different approaches to one another \cite{finch2020towards}. On the other hand, automatic evaluation metrics have proven their utility with their efficiency and undemanding approach to dialogue assessment, but are still considered less reliable in comparison to human judgement \cite{wen2016multi}.  A shortage of task-oriented open-source datasets also acts as a bottleneck in the progression of the field, especially when approaching multiple domains. All of which is compounded by a growing expectation of the average user, as TODS are generally becoming more and more innovative on a global scale.

\section*{Acknowledgement}
This work has been funded by IT Services at UWE-Bristol. Authors would like to acknowledge the assistance provided by Mark Davis and his team in the initial discussions leading to this study and for their collaboration on the project.

\bibliographystyle{elsarticle-num}
\bibliography{performance_TODS}

\end{document}